\documentclass[a4paper,conference,oldfontcommands]{IEEEtran}
\usepackage{color}
\usepackage{lineno}
\usepackage{subfigure}
\usepackage{graphicx}
\usepackage{wrapfig}
\usepackage{amsmath}
\usepackage{caption}
\usepackage{multirow}
\ifCLASSINFOpdf
\else
\fi
\begin{document}
\title{Gait Cycle Reconstruction and Human Identification from Occluded Sequences}
\author{\IEEEauthorblockN{\small{Abhishek Paul}}
\IEEEauthorblockA{\small{Department of ECE}\\
\small{National Institute of Technology,} \\\small{Durgapur, India 713209}\\
\small{Email: ap.19u10127}\\\small{@btech.nitdgp.ac.in}}
\and
\IEEEauthorblockN{\small{Manav Mukesh Jain}}
\IEEEauthorblockA{\small{Department of CSE}\\
\small{Indian Institute of Technology}\\ \small{(BHU), Varanasi, India 221005}\\
\small{Email: manavmukeshjain.cse18}\\\small{@iitbhu.ac.in}}
\and
\IEEEauthorblockN{\small{Jinesh Jain}}
\IEEEauthorblockA{\small{Department of CSE}\\
\small{Indian Institute of Technology}\\ \small{(BHU), Varanasi, India 221005}\\
\small{Email: jineshjain.cse18}\\\small{@iitbhu.ac.in}}
\and
\IEEEauthorblockN{\small{Pratik Chattopadhyay*}}
\IEEEauthorblockA{\small{Department of CSE}\\
\small{Indian Institute of Technology}\\ \small{(BHU), Varanasi, India 221005}\\
\small{Email: pratik.cse}\\\small{@iitbhu.ac.in}\\
\small{Phone: +91-542-716-5325}}\\
}
\maketitle
\begin{abstract}
Gait-based person identification from videos captured at surveillance sites using Computer Vision-based techniques is quite challenging since these walking sequences are usually corrupted with occlusion, and a complete cycle of gait is not always available. In this work, we propose an effective neural network-based model to reconstruct the occluded frames in an input sequence before carrying out gait recognition. Specifically, we employ LSTM networks to predict an embedding for each occluded frame both from the forward and the backward directions, and next fuse the predictions from the two LSTMs by employing a network of residual blocks and convolutional layers. While the LSTMs are trained to minimize the mean-squared loss, the fusion network is trained to optimize the pixel-wise cross-entropy loss between the ground-truth and the reconstructed samples. Evaluation of our approach has been done using synthetically occluded sequences generated from the OU-ISIR LP and CASIA-B data and real-occluded sequences present in the TUM-IITKGP data. The effectiveness of the proposed reconstruction model has been verified through the Dice score and gait-based recognition accuracy using some popular gait recognition methods. Comparative study with existing occlusion handling methods in gait recognition highlights the superiority of our proposed occlusion reconstruction approach over the others. 
\end{abstract}

\IEEEpeerreviewmaketitle

\section{Introduction}\label{intro}
Gait recognition refers to the verification/ identification of individuals from their walking style.  
Person identification from their gait signatures using Computer Vision techniques can be done from a distance without their physical participation and research work in this domain \cite{1561189,S0165168411003392,gupta2021exploiting,gupta2021gait} have shown immense potential of gait to be used as a biometric for identifying suspects in public places. A limitation of these existing approaches is that these work effectively only if at least a clean unoccluded gait cycle of an individual can be captured. However, in real-life situations, occlusion is an inevitable occurrence and to date research on occlusion handling in gait recognition \cite{279851325,Chattopadhyay2014Jan,Chattopadhyay2015Oct,Babaee2018GaitRF,babaee2018person} is not matured enough. The existing techniques are based on certain unrealistic assumptions such as (i) gait features over a cycle must follow a multivariate Gaussian \cite{279851325}, or (ii) multiple occluded gait cycles must be present in a sequence from which a complete cycle can be reconstructed by sampling appropriate frames \cite{Babaee2018GaitRF,babaee2018person}. Also, the work in \cite{Chattopadhyay2014Jan,Chattopadhyay2015Oct} compare gait features based on the matching set of key poses present in both the training and test sequences, which is not effective if a pair of training and test sequences consists of mutually exclusive sets of key poses. There are significant scopes for extending the existing research in this domain to make gait recognition more suitable for application in real-life surveillance sites.

In this work, we consider the problem of occlusion reconstruction in gait sequences and propose a new algorithm to predict the missing/occluded frames in a given sequence using 
Deep Learning models. 
First, a convolutional Autoencoder is used to get an encoded representation of each frame in the sequence. Two LSTMs, namely, $M_1$ and $M_2$ are used to predict the embedding for each missing/occluded frame once from the forward direction using a few preceding frames and once from the backward direction using a few succeeding frames. 
The encoded vectors predicted by $M_1$ and $M_2$ are next decoded using a Decoder network to obtain two image frames corresponding to the same occluded frame. These two frames are combined using a fusion network of residual blocks and convolutional layers to obtain the final reconstructed image corresponding to an occluded frame. 
The main contributions of the work are as follows:
\begin{itemize}
    \item LSTM-based reconstruction of occluded frame embedding from the forward and backward directions
    
    \item Fusing the decoded predictions from two LSTMs using a network of residual blocks and convolution layers
    
    \item Making the pre-trained models publicly available to the research community for further comparative studies
\end{itemize}

\section{Related work}\label{rw}
The first work on Computer Vision-based gait recognition as given in \cite{1561189} computes a feature termed the Gait Energy Image (GEI) by averaging frame-level features over a gait cycle. Due to averaging features, the GEI fails to capture the dynamics of gait appropriately. Some extensions to this feature have been proposed in the later years. 
For example, the work in \cite{S0165168411003392} describes a feature termed the Pose Energy Image which is essentially a collection of GEI features corresponding to several fractional parts of a gait cycle that captures the dynamics of gait at a higher resolution. Other similar approaches in this category include that in 
\cite{Chattopadhyay2014Jan} and \cite{Chattopadhyay2015Oct}. Another improvement over the traditional GEI feature is described in \cite{Zhang2010Jul} in which the active walking regions are computed by subtracting adjacent binary silhouette frames and next these difference images are aggregated to compute a feature termed the Active Energy Image. 
The work in \cite{gupta2021exploiting} is an improvement over \cite{S0165168411003392} in which instead of considering a single set of key poses, multiple key pose sets of varying lengths are considered and gait features are constructed and compared with the gallery set after mapping the frames to each set of key poses. The approach discussed in \cite{gupta2021gait} utilizes a Pix2Pix GAN to predict silhouettes without covariate objects from binary silhouette sequences with covariate objects and uses a pose-based feature termed the Dynamic Gait Energy Image for individual recognition.

Among the Deep Learning-based approaches, 
in \cite{8835194}, 
the  dynamic parameters of motion are computed and fed into a Convolutional Neural Network 
with four convolutional layers, four sub-sampling layers, one fully-connected layer, and a final softmax layer for person identification. 
In \cite{Shiraga2016Jun}, the GEI features are passed through a CNN termed the GEINet to obtain deep gait representations for person identification. In another work  \cite{Alotaibi2017Nov}, a small-scale CNN with four convolutional layers and four pooling layers is considered for performing deep feature-based gait recognition. The work in \cite{Battistone2019Sep} leverages structured data and temporal information by employing a spatio-temporal deep LSTM neural network model. This approach helps in learning long and short-term dependencies in the gait sequences with a graph structure. 
Some popular approaches towards view-invariant gait recognition include the one in \cite{8616800} in which a low-dimensional
discriminant subspace is learned that effectively links the GEI features across multiple views,  and \cite{chao2019gaitset,chao2021gaitset} in each of which the GaitSet model is used to extract useful spatio-temporal information from given sequences and integrate this information to improve the recognition accuracy. The GaitPart model given in \cite{fan2020gaitpart} is an improvement over GaitSet which consists of a frame-level part feature extractor that provides an intrinsic representation of the micro-motions at the different body parts followed by a temporal feature aggregator. A distilled version of the GaitSet model is proposed in \cite{song2022distilled} in which a lightweight student CNN model is designed and trained using a joint knowledge distillation algorithm so that the features extracted by the student model are similar to that by the more complex teacher GaitSet model. Some other approaches in this category include \cite{yu2017gaitgan} that employs a GAN to translate the GEI feature captured from one view to another view, and an improved version of the same algorithm \cite{yu2019gaitganv2} that incorporates multi-loss training. The work in \cite{he2018multi} also uses a GAN to learn view-specific features  along with temporal information through a feature termed the Period Energy Image. A view normalizing conditional GAN is presented in \cite{zhang2019vn} to learn the identity-related representations from multiple views. In \cite{zhang2019vt}, a StarGAN model has been used for multi-view transformation from a single view. More recently, in \cite{han2022unified} a view-invariant gait recognition approach is presented in which an angular softmax loss function is used to learn separable features in the Cosine space, and simultaneously another triplet loss function is used to increase the separation between the feature vectors from different subjects. Finally, a batch-normalization layer is employed to effectively optimize the two loss terms.

The above techniques require a complete gait cycle for proper functioning 
and are not expected to perform well in the presence of occlusion where a complete gait cycle may not be available.  
A few approaches toward occlusion handling in gait recognition are discussed next. 
In \cite{279851325}, recognition is performed after reconstructing the missing/occluded frames using a Gaussian Process Dynamic Model. 
This method is based on the assumption that the gait features over a cycle follow a multi-variate Gaussian, which is not expected to be true always. 
The work in \cite{Chattopadhyay2015Oct,chattopadhyay2014frontal} use Kinect-captured depth, RGB, and skeleton streams for video-based gait recognition in the presence of occlusion. These methods attempt to extract gait features and perform classification using only the available information from the unoccluded frames present in the input sequence and do not reconstruct the missing frames. Among these, in \cite{chattopadhyay2014frontal} Kinect RGB-D streams are used to develop a three-step hierarchical classification approach in which the following features are used at the different hierarchical levels: (i) soft biometric features, (ii) motion-level features derived from joint trajectories, and (iii) silhouette-level features.
Comparison at each level is done based on the available fractions of gait cycle present in both the training and test sequences. The work in \cite{Chattopadhyay2015Oct} fuses depth-based silhouette features from the back view and skeleton-based features corresponding to the lower body from the front view and follows a key pose-based matching approach for classification similar to that used in \cite{Chattopadhyay2015Oct}. In \cite{babaee2018person}, a deep neural network-based generative network has been used to predict the GEI corresponding to a complete cycle by taking as input the GEI computed from the available frames of an incomplete cycle. 
The performance of this approach is also highly dependent on the quality of the input GEI computed from the incomplete cycle, since a moderate to high degree of occlusion is expected
to degrade this feature, making it difficult for the network to predict the final GEI accurately. 

From the extensive literature survey, we observe that 
significant attention has been given to developing neural network-based approaches for reconstructing occluded/missing frames in a gait sequence. Since gait follows a spatio-temporal pattern, and LSTMs are popularly used to deal with sequential data, it appears that the use of an LSTM-based generator would help in accurately predicting the silhouette features of an occluded frame using the information from a few preceding or succeeding frames, which we study in this work. 
\section{Proposed approach} \label{oa}
\begin{figure}[ht]
\scriptsize
\centering
\includegraphics[width=0.45\textwidth]{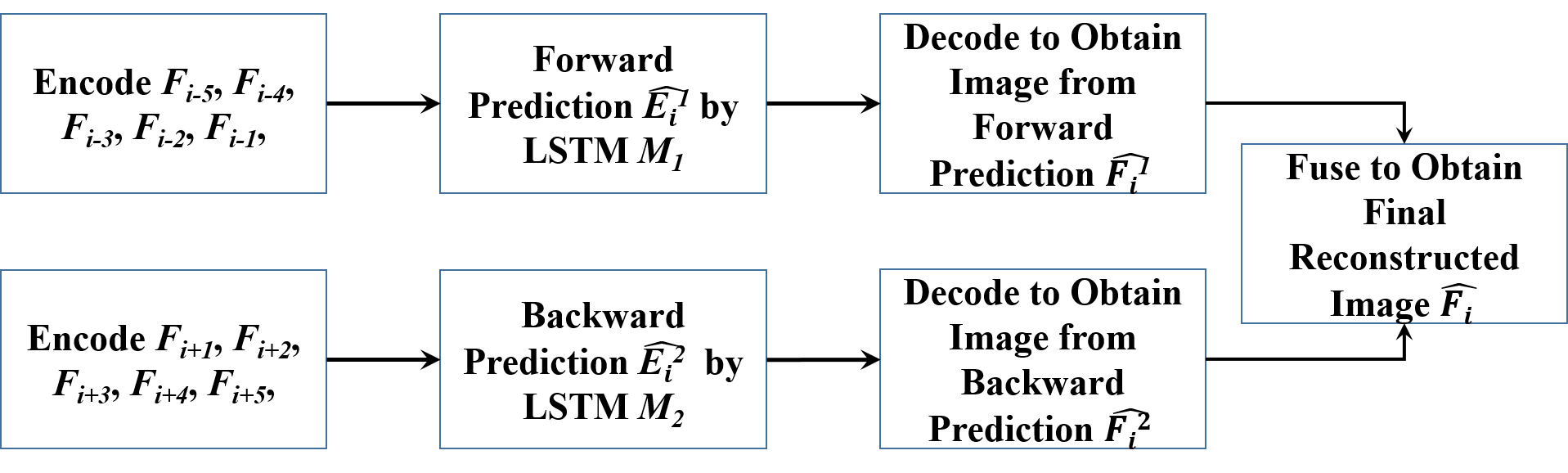}
\caption{A schematic diagram showing reconstruction of an occluded frame $F_i$ from its preceding and succeeding five unoccluded frames}\label{bd}
\end{figure}

\begin{figure}[!ht]
\centerline{\includegraphics[height = 0.9\textheight, width = \linewidth]{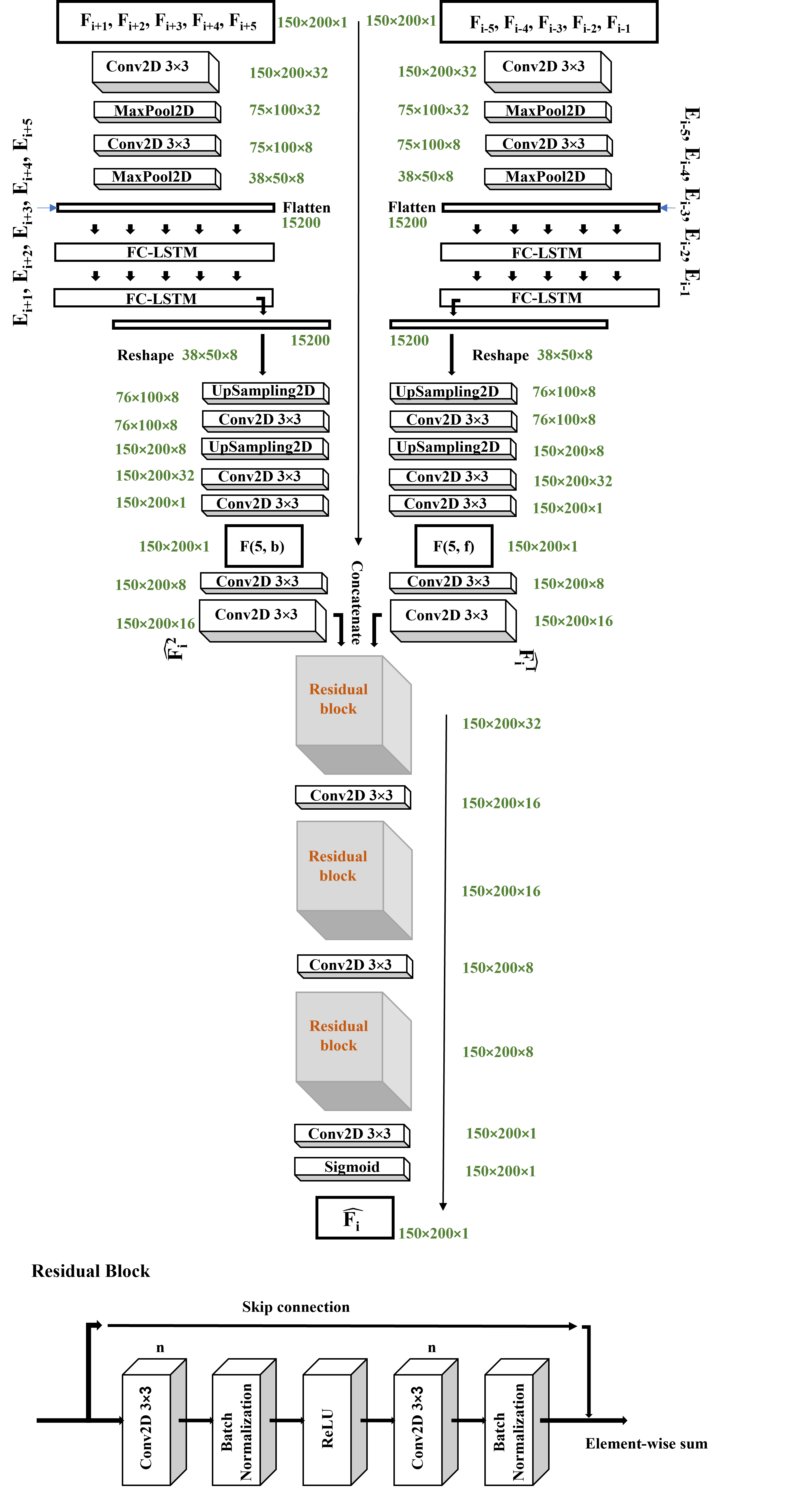}}
\caption{Architecture of the proposed occlusion reconstruction model consisting of Encoder, pair of LSTMs, Decoder, and fusion network}\label{net}
\end{figure}

As in any appearance-based gait recognition approach \cite{1561189,S0165168411003392,gupta2021exploiting,gupta2021gait}, here also we perform reconstruction and recognition using normalized binary silhouettes extracted from RGB frames. 
The main objective of the work is to develop a neural model for the reconstruction of occluded frames in a gait sequence. Our proposed model is a fusion of multiple sub-networks as follows: (i) an Encoder that encodes the binary silhouette frames, (ii) two LSTMs $M_1$ and $M_2$ to predict encoded representations of the binary silhouettes corresponding to a missing/occluded frame from the forward and the backward directions using five preceding and five succeeding frames, respectively, (iii) a Decoder network to transform the reconstructed encoded frames back to the image space, and finally (iv) a fusion model to combine the predictions from the two LSTMs.
A schematic diagram explaining the steps of our occlusion reconstruction algorithm is shown in Figure \ref{bd} and 
the complete architecture of the proposed occlusion reconstruction model is shown in Figure \ref{net}. 
With reference to this figure, suppose we have a sequence of $N$ preprocessed binary frames $F_1$, $F_2$, ..., $F_N$, each of dimension 150$\times$200, out of which a few frames are occluded. To reconstruct the silhouette corresponding to an occluded frame $F_i$, we make a forward prediction $\hat{F^1_i}$ through $M_1$ by using the spatio-temporal information given by five previous frames, i.e., $F_{i-5}$, $F_{i-4}$, $F_{i-3}$, $F_{i-2}$, $F_{i-1}$, and a backward prediction $\hat{F^2_i}$ through $M_2$ using the spatio-temporal information corresponding to the five succeeding frames, i.e., $F_{i+1}$, $F_{i+2}$, $F_{i+3}$, $F_{i+4}$, $F_{i+5}$. While using $M_1$ to make the forward prediction, the frames preceding $F_{i}$ must be either unoccluded or predicted at a previous step. Similarly, while making the backward prediction using $M_2$, the frames succeeding $F_i$ must be either unoccluded or already predicted previously. It may be noted that the predictions $\hat{F^1_i}$ and $\hat{F^2_i}$ are independent of whether frame $F_i$ is partially or completely occluded, since information about the frame $F_i$ has not been used to make the prediction by either of the LSTMs $M_1$ and $M_2$.
Instead of carrying out the computations in the image space, we propose to do the same in a low-dimensional encoded space to reduce the adverse impact of noise and other redundant image features on the prediction and also to make the process time-efficient. 

A two-layer Convolutional Autoencoder has been used here to obtain an embedding for each frame. With reference to Fig. \ref{net}, the Encoder network of the Convolutional Autoencoder consists of two convolutional layers, each followed by a max-pooling layer. Each frame $F_j$ used to reconstruct the occluded frame $F_i$ is passed through the Autoencoder to get an encoded representation, say $E_j$. During the forward prediction phase, after passing each frame, one at a time, through the first set of convolutional and max-pooling layers, we obtain a 75$\times$100$\times$32 dimensional activation map. Next, after passing the above features through the second set of convolutional and max-pooling layers, a 38$\times$50$\times$8 dimensional feature map is obtained. This feature map is flattened into a 15200-dimensional feature vector before inputting into the LSTM-based occlusion reconstruction model. To predict the occluded frame $F_i$, we compute two sets of embedding: (i) from the previous five frames denoted by $E_{i-5}$, $E_{i-4}$, $E_{i-3}$, $E_{i-2}$, $E_{i-1}$, and (ii) from the subsequent five frames denoted by $E_{i+1}$, $E_{i+2}$, $E_{i+3}$, $E_{i+4}$, $E_{i+5}$.

\begin{figure}[ht]
\begin{center}
\includegraphics[height = 0.15\textheight]{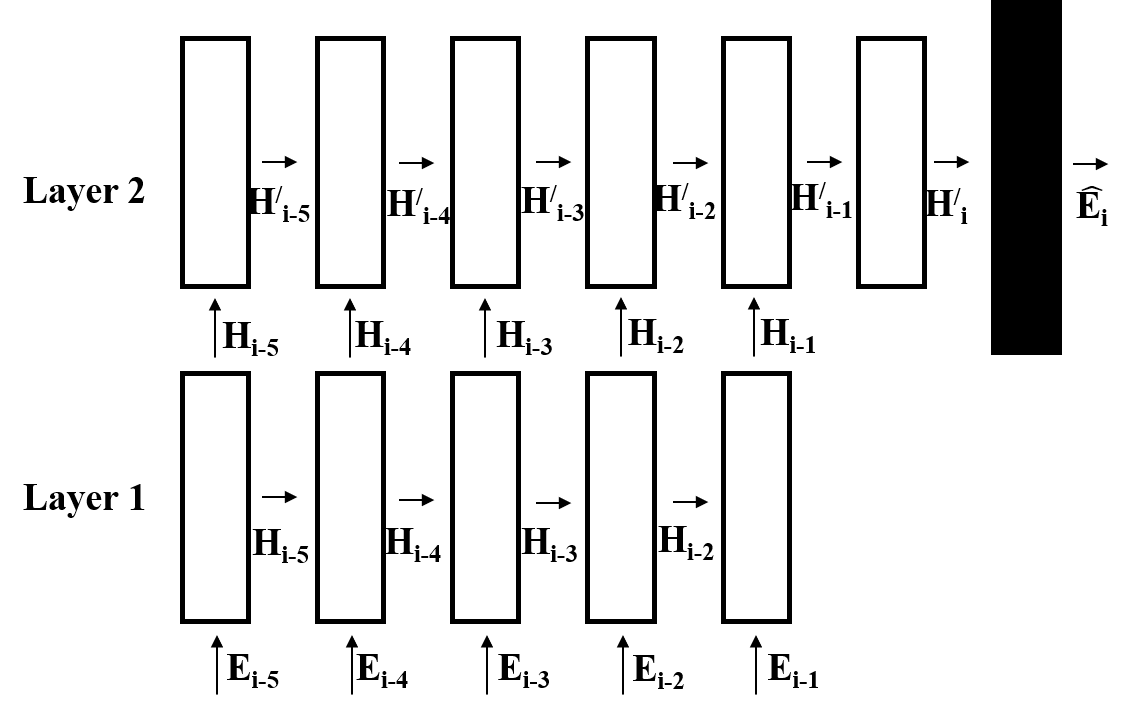}
\end{center}
\caption{Schematic diagram of the LSTM architecture used in our work for forward prediction}\label{lstm}
\end{figure}

The forward LSTM $M_1$ uses the first set of embedded vectors to predict $\hat{E^1_i}$, i.e., a first estimate of the embedding for the occluded frame $F_i$, whereas the backward LSTM $M_2$ uses the second set of embedded vectors to predict $\hat{E^2_i}$, i.e., a second estimate of the same occluded frame $F_i$. An overview of the LSTM architecture used in our work for the forward prediction, i.e., LSTM $M_1$, is shown in Fig. \ref{lstm}. 
As shown in the figure, the LSTM $M_1$ consists of two stacked layers of five cells each. For prediction of embedding $\hat{E_i^1}$ corresponding to frame $F_i$, the inputs to each of the five cells of LSTM $M_1$ in the first layer are an embedded vector output by the Encoder along with the hidden layer features output by the previous LSTM cell. On the other hand, the inputs to each cell in the second layer are the outputs from the corresponding cell in the first layer and the output from the previous cell in the same layer. The five cells in the first layer output five 1024-dimensional feature vectors denoted by $H_{i-5}$, $H_{i-4}$, $H_{i-3}$, $H_{i-2}$, and $H_{i-1}$. The LSTM layer stacked on top of this first layer exploits the spatio-temporal information contained in the sequence by combining the output vectors from the first layer to generate a final 1024-dimensional vector $H_i'$ which is further passed through a dense layer with 15200 neurons to obtain the desired predicted embedding for frame $F_i$, i.e., $\hat{E^1_i}$. A similar architecture is also used for backward prediction via $M_2$ to obtain a second prediction for frame $F_i$, denoted by $\hat{E^2_i}$. 
Mathematically, if $T_f$ and $T_b$ denote the functions learned by the LSTMs $M_1$ and $M_2$, then the predicted embedding vectors $\hat{E^1_i}$ and $\hat{E^2_i}$ for any occluded frame $F_i$ are obtained as:
\begin{eqnarray}
    \hat{E^1_i} &=& T_f (E_{i-5}, E_{i-4}, E_{i-3}, E_{i-2}, E_{i-1}),\\
    \hat{E^2_i} &=& T_b (E_{i+1}, E_{i+2}, E_{i+3}, E_{i+4}, E_{i+5}).
\end{eqnarray}
Each of $\hat{E^1_i}$ and $\hat{E^2_i}$ is next passed through a Decoder network to obtain predictions $\hat{F^1_i}$ and $\hat{F^2_i}$ for frame $F_i$ in the image space. This Decoder network is the mirror-twin of the Encoder. It consists of sets of up-sampling layers followed by convolutional layers. The LSTM-generated embedding is reshaped into a 38$\times$50$\times$8 dimensional feature map which is further transformed to a 76$\times$100$\times$8 dimensional feature map through up-sampling and further passing through a convolutional layer with eight 3$\times$3 filters. The output from this layer set is passed through another identical set of up-sampling and convolutional layers consisting of 32 filters yielding a 152$\times$200$\times$32 dimensional feature map. To match the dimensions of the generated feature map with the desired image dimensions, i.e., 150$\times$200, strips of width one pixel are removed from either side of the feature maps along the height to make its dimension 150$\times$200$\times$32. The 32 feature maps are finally compressed into the desired reconstructed image of dimensions 150$\times$200$\times$1 
by passing through another convolutional layer with a single filter. 

During the training phase, the Convolutional Autoencoder (formed by stacking the Encoder and Decoder), and the two LSTMs $M_1$ and $M_2$,  are trained separately. If the input to the Autoencoder is a binary frame $F_i$, and the output generated by the Autoencoder for this particular input is $\hat{F_i}$, then the reconstruction loss ($L_{rec}$) used to train the model is computed as the pixel-wise binary cross-entropy loss and is given by:
\begin{equation}\label{rec}
    L_{rec}\!\!=\!\! \sum_{i=1}^{\mathcal{N}}\!\sum_{p \in P}\!- F_{i}(p) log(\hat{F_{i}}(p)) \!-\! (1\!-\!F_{i}(p)) log(1\!-\!\hat{F_{i}}(p)).
\end{equation}
Here, $P$ denotes the set of pixels in a frame, $F_i(p)$ and $\hat{F_i}(p)$ respectively denote the intensity of the $p^{th}$ pixel in ground truth frame $F_i$ and predicted frame $\hat{F_i}$, and $\mathcal{N}$ denotes the number of images in a batch used to train the Autoencoder. 

\begin{figure}[ht]
\centering
\includegraphics[width = 0.47\textwidth]{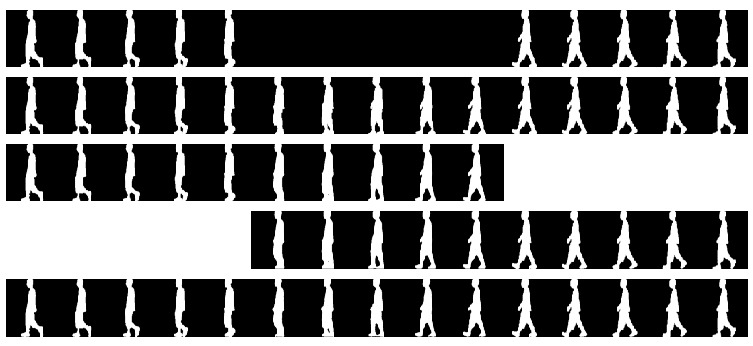}
\caption{$1^{st}$ row: Synthetically Occluded sequence, $2^{nd}$ row: Ground Truth, $3^{rd}$ row: Sequence reconstructed by $M_1$, $4^{th}$ row: Sequence reconstructed by $M_2$, $5^{th}$ row: Sequence reconstructed by the proposed fusion network} \label{preds}
\end{figure}

The predicted feature maps $\hat{F^1_i}$ and $\hat{F^2_i}$ by $M_1$ and $M_2$ have intensity values in the range \textit{0} to \textit{1}. These are first binarized using 0.5 as the threshold and next passed through the fusion network that consists of alternate residual blocks (with convolutional layers followed by batch normalization) and convolutional layers with 3$\times$3 filters as shown in Fig. \ref{net}. This network is trained by employing the Adam optimizer to minimize the binary cross-entropy loss between the LSTM-generated images and the corresponding ground truth image. 
The first row in Fig. \ref{preds} shows a sample binary silhouette sequence with 15 frames in which a few frames are occluded/missing while the second row shows the corresponding ground truth frames. The missing frames predicted by the LSTMs $M_1$ and $M_2$ are shown in the third and fourth rows, and the  reconstructed sequence obtained from the fusion network is shown in the fifth row of Fig. \ref{preds}. 
Once the reconstructed frames are obtained any standard gait recognition method can be employed to identify an individual. 

\section{Experimental evaluation} \label{ee}
All experiments have been carried out on a system with 96 GB RAM, one i9-18 core processor, along with three GPUs, namely, one Titan Xp with 12 GB  RAM, 12 GB  frame-buffer  memory,  and 256 MB  BAR1  memory and, two  GeForce  GTX1080Ti  with 11 GB  RAM, 11 GB  frame-buffer memory, and 256 MB BAR1 memory. 

\underline{Training the Individual Sub-Networks}: 
The sub-networks that build up the proposed occlusion reconstruction model must be trained using an extensive gallery set in a manner such that the missing/occluded frames in any walking sequence can be accurately predicted. On the other hand, the gait recognition model is a discriminative model that must be trained with a specific gallery set corresponding to a particular dataset. Next, we provide details of the training methodology for each of the individual sub-networks used in the reconstruction model.


\textit{Autoencoder}: The Autoencoder network is trained on four random hand-picked gait sequences from each of the CASIA-B \cite{zheng2011robust} and the OU-ISIR Large Population (LP) \cite{6215042} gait datasets, resulting in a total of 224 silhouettes covering all types of body poses. The model has been trained for 100 epochs or till saturation, whichever is earlier. The network is optimized using the Adam optimizer with a learning rate of 0.001 and an exponential decay rate 
of 0.9. 
    
\textit{Forward and Backward LSTMs}: These networks have been trained on the encoded data output by the encoder corresponding to the sequences from the CASIA-B and the OU-ISIR LP datasets. Specifically, we train LSTM $M_1$ with 184 sets of embedding from five consecutive frames as input and embedding from the sixth frame as the desired ground truth. Similarly, we train LSTM $M_2$ with the embedding from the first frame as the desired output and the embedding corresponding to the succeeding five frames as input. Both $M_1$ and $M_2$ are trained by optimizing the mean squared error loss through the Adam optimizer with a learning rate of 0.001, the exponential decay rates for the first moment estimates being 0.9 and that for the second moment estimates being 0.999.

\textit{Fusion network}: The images predicted by the LSTMs $M_1$ and $M_2$ from the above sequences along with the corresponding ground truth form the training set for this model. 

\underline{Experiments and Results}: 
We evaluate the effectiveness of the occlusion reconstruction model using synthetically occluded sequences generated from the CASIA-B \cite{zheng2011robust} and the OU-ISIR LP \cite{6215042} datasets, and real occluded sequences from the TUM-IITKGP dataset \cite{hofmann2011gait}. Specifically, for the CASIA-B data we consider the normal walking sequences from all the 124 subjects present in this data, and for the OU-ISIR LP we consider normal walking sequences from 80 randomly selected subjects present in this dataset. 
Corresponding to each of the two datasets, we consider two complete gait cycles as training sets for gait recognition, and four gait cycles for testing after corrupting the sequences with synthetic occlusion of varying amounts.
It may be noted that none of these test sequences forms part of the gallery set for training either the sub-networks of the occlusion reconstruction model or the gait recognition model. 
The synthetically occluded test sequences are generated by randomly selecting a few frames in these sequences and blackening them, i.e., removing the silhouette information from those frames completely. For the TUM-IITKGP data, we consider four unoccluded gait cycles corresponding to each of the 35 subjects to train the gait recognition model and two occluded sequences for each subject (one for each of static and dynamic occlusions) to form the test set. While working with the TUM-IITKGP data, we first employ a VGG-16 model to automatically detect the occluded frames present in a sequence using an approach similar to that given in \cite{das2019rgait} and next use the proposed model to reconstruct the occluded frames. 

\begin{figure}[ht]
\centering
\includegraphics[height = 0.13\textheight]{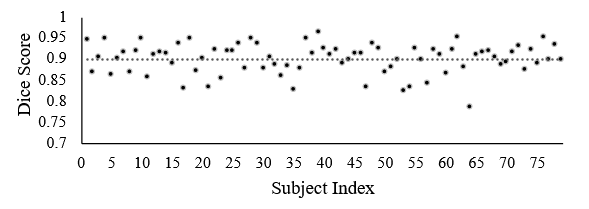}
\caption{Dice scores between the GEIs of reconstructed and original sequences for 80 subjects}\label{dice}
\end{figure}

In our first experiment, we quantitatively evaluate the effectiveness of the proposed reconstruction model in terms of similarity of the GEI features computed from the reconstructed sequence with the GEI features constructed from the original unoccluded sequence. For this, 
we form a test set by considering 80 distinct gait cycles from the test sequences of CASIA-B and OU-ISIR LP data and synthetically occlude (i.e., blacken) the frames in these cycles by varying the occlusion degree between 0-50\% in each of these gait cycles. Next, the trained reconstruction model is used to predict the missing/occluded frames in each sequence, following which the GEI features are computed from the reconstructed sequences. 
The sequences from the TUM-IITKGP data could not be used in this experiment since it consists of sequences with real occlusion and there are no ground truth unoccluded frames available corresponding to the occluded frames that can be used to compare with the reconstructed frames. A similarity score between the original GEI and the reconstructed GEI is henceforth computed based on the Dice Similarity Coefficient score \cite{carass2020evaluating} (in short \textit{Dice score}). The plot in Figure \ref{dice} shows the individual Dice scores for the 80 gait cycles used in this experiment. 
The Dice score ranges from `0' to `1', with `0' and `1' implying \textit{no similarity} and \textit{complete similarity}, respectively.  It is observed from the figure that the Dice score value between the GEIs computed from the reconstructed frames and the ground truth frames for each subject varies between 0.80 and 0.95, with an average 
close to 0.90 which is quite high, emphasizing the effectiveness of the proposed occlusion reconstruction model. 

Next, the performance of the reconstruction model is tested through gait recognition accuracy on the reconstructed sequences for both the OU-ISIR LP and CASIA-B datasets. As gait feature we consider the GEI due to its simplicity and as a classification model, we consider a Random Forest (with Bagging) consisting of 100 decision trees.

The Random Forest is trained using two GEIs per subject present in the training set. For testing, we synthetically occlude (i.e., blacken) each cycle present in the test set by considering varying degrees of occlusion, namely, 5-10\%, 10-20\%, 20-30\%, 30-40\%, and 40-50\%.
For each degree of occlusion, we compute the Dice score of reconstruction and the Rank 1 recognition accuracy and present the results 
in Table \ref{tab1}. 

\begin{table}[ht]
\scriptsize
\begin{center}
\caption{Reconstruction Dice score and Rank-1 accuracy on the test sets for (a) OU-ISIR LP and (b) CASIA-B datasets}\label{tab1}
\begin{tabular}{ |c|c|c|c|c| } 
\hline
\textbf{Percentage of} & \multicolumn{2}{c|}{\textbf{OU-ISIR LP}} & \multicolumn{2}{c|}{\textbf{CASIA-B}} \\
\cline{2-5}
\textbf{Occlusion}&\textbf{Dice score}&\textbf{Accuracy (\%)}&\textbf{Dice score}&\textbf{Accuracy (\%)}\\
\hline
5-10\% & 0.96 &97.50 &0.99& 92.40 \\
10-20\% & 0.90 &91.25 &0.96& 84.81 \\
20-30\% &0.87 &78.75 &0.87& 73.41 \\
30-40\% &0.82 &63.75 &0.87& 72.15 \\
40-50\% &0.78 &51.25 &0.80& 63.29 \\
\hline
\end{tabular}
\end{center}
\end{table}
As expected, it can be seen from the table that the Dice score values for both the synthetically occluded datasets are quite high (i.e., close to \textit{1}) if the degree of occlusion is low. For higher degrees of occlusion, the Dice score reduces slightly due to the propagation of errors in the previously reconstructed frames. Still, a Dice score value close to 0.80 for heavily occluded sequences (i.e., for 40-50\% occlusion) can be considered to be quite good. 
\color{black}
It is also seen from the table that the gait recognition accuracy values for both the datasets are quite high if the percentage of occlusion is within 20\%, which in turn highlights the effectiveness of our reconstruction model. 
Also, the Rank 1 accuracy achieved by our proposed method under heavy occlusion situations, i.e., 40-50 \% occlusion is above 50\%, which can be said to be reasonably good. It may be noted that the above accuracy values are computed based on the GEI feature and this performance is expected to improve further by using more sophisticated gait features.

\begin{figure}[ht]
\centering
    \begin{subfigure}[ ]{
    \includegraphics[height = 0.20\textheight]{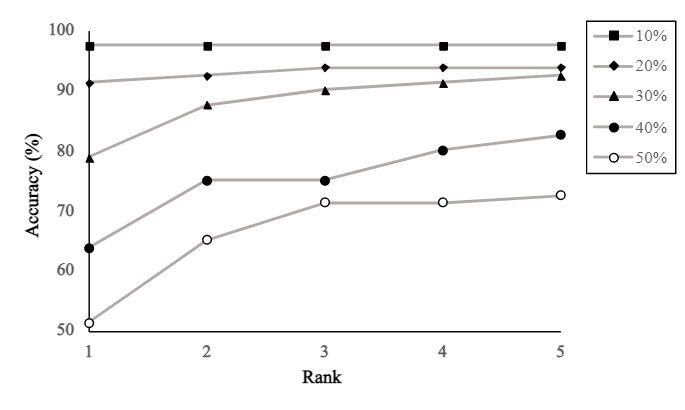}}
    \end{subfigure}\\
    
    \begin{subfigure}[ ]{
    \includegraphics[height = 0.20\textheight]{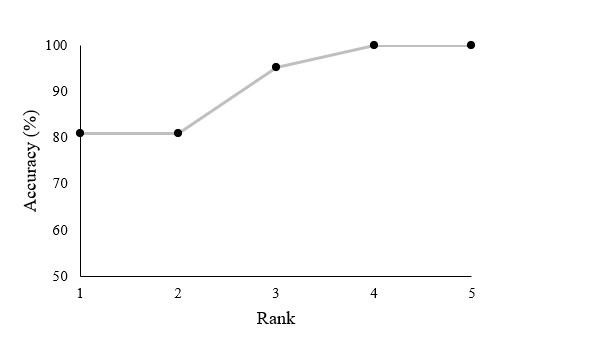}}
    \end{subfigure}
    \caption{Rank-based accuracy through CMC curves on (a) synthetically occluded sequences of OU-ISIR LP data for varying degrees of occlusion, (b) real occluded sequences of TUM-IITKGP data}\label{ou-rank}
\end{figure}

Next, we study the rank-wise improvement in the gait recognition accuracy on synthetically occluded test sequences of the OU-ISIR LP data and the real occluded sequences of the TUM-IITKGP data through Cumulative Match Characteristic (CMC) curves. The same training and test sets for each dataset as discussed have also been used here. Corresponding results are shown in Figs. \ref{ou-rank}(a) and (b) for the two datasets as the value of the 
rank is increased from 1 to 5. 
It is observed from the plot in Fig. \ref{ou-rank} that for the synthetically occluded OU-ISIR LP data, the accuracy values for the different ranks are quite high if the percentage of occlusion is below 30\%. For moderately high 20-30\% occlusion also, we obtain more than 85\% accuracy at Rank 2. As expected, the accuracy at the different rank values deteriorates once the percentage of occluded frames in gait cycles is increased beyond 40\%, but still the recognition performance is reasonably good even for such high degrees of occlusion. Even for real occluded sequences of the TUM-IITKGP data, the recognition accuracy after reconstruction is 80\% at Rank 1 which increases to 100\% at Rank 4.  

\color{black}
Finally, we make a comparative performance analysis of our proposed approach with other popular gait recognition techniques existing in the literature. 

Among the important occlusion handling methods in gait recognition, the work in \cite{Babaee2018GaitRF,babaee2018person} predict the GEI features of an individual by taking as input the GEI corresponding to an occluded sequence of the same person and do not carry out frame-level reconstruction. In contrast, the method in \cite{royocclusion} performs frame-level reconstruction by assuming that the gait features over a cycle follow a Gaussian. The work in \cite{article1} computes gait features from difference images without carrying out any reconstruction process. To compare the different approaches in terms of the effectiveness of reconstruction, we consider the synthetically occluded test sequences of the CASIA-B data corrupted by varying levels of synthetic occlusion randomly between 0\% to 50\% and compute the Dice score between the GEIs derived from the reconstructed sequences and that from the ground-truth unoccluded frames and present the results in Table \ref{tab_comp_dice}. 
\begin{table}[ht]
\scriptsize
\centering
\caption{Comparative study of the reconstruction quality of the different gait recognition approaches in terms of Average Dice Score between generated and ground-truth GEIs on the synthetically occluded test set generated from the CASIA-B data}\label{tab_comp_dice}
\begin{tabular}{|c|c|c|c|c|} 
\hline
\textbf{Approach}&\cite{Babaee2018GaitRF}&\cite{babaee2018person}&\cite{royocclusion}&Ours\\
\hline
\textbf{Dice Score}&0.88&0.92&0.70&0.95\\
\hline
\end{tabular}
\end{table}
It can be seen from the table that our proposed fusion model performs reconstruction better than each of the other compared methods in terms of Dice score. 

Next, we compare the Rank 1 recognition accuracy given by the above-mentioned occlusion handling methods as well as that given by the following non-occlusion handling methods: \cite{1561189,S0165168411003392,Zhang2010Jul,Shiraga2016Jun,Alotaibi2017Nov,gupta2021exploiting,gupta2021gait} on the reconstructed sequences generated by our reconstruction model. We consider the same synthetically occluded test set constructed from the CASIA-B data and also the real occluded sequences from the TUM-IITKGP data. 
The same gallery sets for training the recognition model corresponding to the CASIA-B and the TUM-IITKGP data as used in the previous experiments have also been used here to compute the gait recognition accuracy. Corresponding results are shown in 
Table \ref{tab_comp}. 

\begin{table}[ht]
\scriptsize
\centering
\caption{Comparative study of different gait recognition approaches in terms of Rank-1 accuracy on real occluded sequences of TUM-IITKGP data and synthetically occluded test set generated from the CASIA-B data\\}\label{tab_comp}
\begin{tabular}{ |c|c|c|} 
\hline
\textbf{Approach} & \textbf{TUM-IITKGP (\%)} & \textbf{CASIA-B (\%)}\\
\hline
 \cite{Babaee2018GaitRF} &72.86&78.92 \\
 \cite{babaee2018person} &77.14&80.00 \\
 \cite{article1} &71.43&77.65 \\
 \cite{royocclusion}&68.57&75.23\\
Our Model + \cite{1561189} &78.57& 80.23 \\
Our Model + \cite{S0165168411003392} &80.00& 83.54 \\
Our Model + \cite{Zhang2010Jul} &78.57& 86.42 \\
Our Model + \cite{Shiraga2016Jun} &78.57& 86.79\\
Our Model + \cite{Alotaibi2017Nov} &81.43& 83.54 \\
Our Model + \cite{gupta2021exploiting} &84.29& 96.37\\
Our Model + \cite{gupta2021gait} &85.71& 95.56\\
\hline
\end{tabular}
\end{table}
It can be seen from the table that for both the datasets, quite accurate results are obtained on fusing our proposed occlusion reconstruction model with gait recognition methods developed for handling unoccluded gait cycles, namely, \cite{1561189,S0165168411003392,Zhang2010Jul,Shiraga2016Jun,Alotaibi2017Nov,gupta2021exploiting,gupta2021gait}. Further, the accuracy by each of these methods always surpasses the accuracy given by the occlusion handling methods in gait recognition, i.e., \cite{Babaee2018GaitRF,babaee2018person,article1,royocclusion}. It is also seen from the table that the recent approaches given in \cite{gupta2021exploiting,gupta2021gait} perform with significantly higher accuracy compared to the other non-occlusion handling methods for both the datasets due to using more sophisticated gait features for recognition. In general, the accuracy values corresponding to the different methods are lower for the TUM-IITKGP data as compared to CASIA-B data. A possible reason for this observation is that the binary silhouettes present in the TUM-IITKGP data are noisier compared to that of the CASIA-B data.

\begin{figure}[ht]
    \centering
    \begin{subfigure}[]{
    \includegraphics[height = 0.20\textheight]{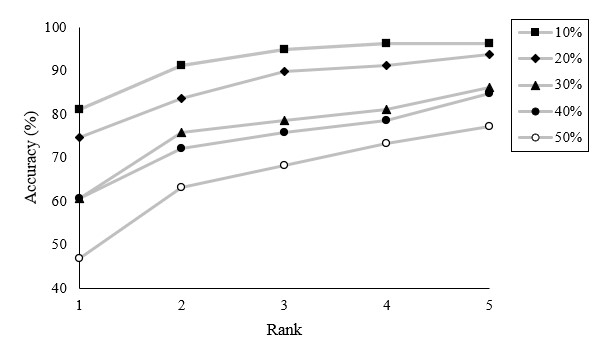}}
    \end{subfigure}
    \begin{subfigure}[]{
    \includegraphics[height = 0.20\textheight]{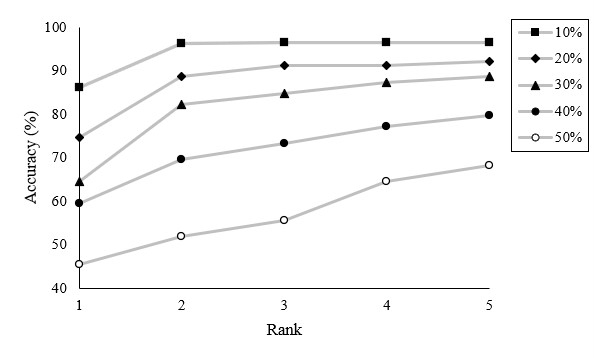}}
    \end{subfigure}
    \begin{subfigure}[]{
    \includegraphics[height = 0.20\textheight]{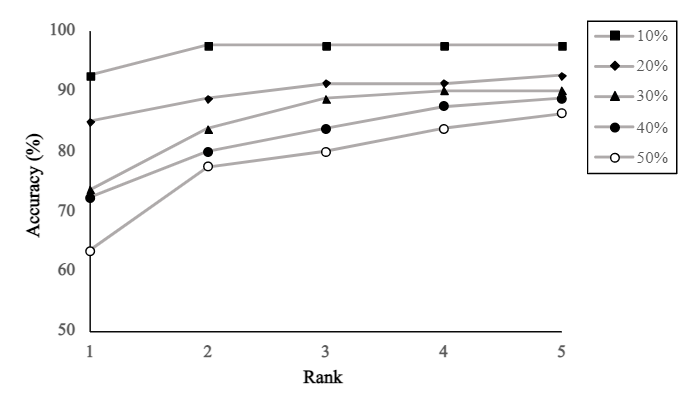}}
    \end{subfigure}
    \caption{Accuracy on test sequences of CASIA-B data for varying degrees of synthetic occlusion consider the following reconstruction models: (a) $M_1$, (b) $M_2$, (c) proposed fused reconstruction model}
    \label{fig:cmc}
\end{figure}

\underline{Ablation Study}: The core reconstruction model proposed in this work is formed by two LSTMs, namely a forward LSTM $M_1$, a backward LSTM $M_2$, and a fusion network that combines the predictions from $M_1$ and $M_2$. In our final experiment, we make an ablation study to observe how effective the individual LSTMs and the proposed fusion network are in reconstructing the occluded frames present in a given input sequence. The test sequences of the CASIA-B data corrupted with varying levels of synthetic occlusion (namely, 10\%, 20\%, 30\%, 40\%, and 50\%) have been used to study the effectiveness of reconstruction through the Dice score metric for $M_1$, $M_2$, and the fused model, separately. The average Dice scores obtained from the three networks by comparing the reconstruction frames with the ground truth for the varying degrees of occlusion are 0.82, 0.87, and 0.90, respectively. We further study the rank-wise improvement in recognition accuracy after reconstructing the above occluded sequences separately by each of the three models and present the corresponding CMC curves in Figs. \ref{fig:cmc}(a)-(c). As recognition model, we use a Random Forest with 100 trees.
The figures show that significantly higher 
recognition accuracy values are obtained at the different ranks corresponding to each of the different degrees of occlusion if sequences reconstructed by the fused model are used for recognition. This model is combination with the Random Forest provides a Rank 5 accuracy of 86\% even for very high degree of occlusion (i.e., 50\%). In contrast, the accuracy obtained using the reconstructed sequences from $M_1$ and $M_2$ are 77\% and 68\%, respectively. The above results based on Dice scores and rank-based recognition accuracy also verify the effectiveness of the proposed fused model.  

\color{black}
\section{Conclusions and Future Work} In this work, we propose a novel occlusion reconstruction model by stacking an Encoder-Decoder architecture with Forward and Backward LSTMs, and a Fusion network. 
The effectiveness of our proposed model is independent of whether an input sequence contains partial or full-body occlusions since while reconstructing an occluded frame, each of the forward and backward LSTM utilizes the spatio-temporal information given by a set of preceding or succeeding consecutive unoccluded frames, that does not include the occluded frame. 
Finally, these two predictions are fused using a set of residual blocks and convolutional layers to come up with a refined prediction about the occluded frame. Experimental results through Dice scores and recognition accuracy on synthetically occluded CASIA-B and OU-ISIR LP data and real occluded TUM-IITKGP data reveal that our model performs occlusion reconstruction quite effectively. \color{black} Impressive results are obtained by combining our reconstruction model with seven popular gait recognition methods developed specifically for handling unoccluded sequences. 
Each of these combinations has been seen to provide a higher recognition rate than the existing occlusion handling methods in gait recognition on occluded datasets. In the future, our approach needs to be validated using extensive real occluded gait datasets  considering other challenging scenarios such as viewpoint changes and different lighting conditions in the gallery and test sequences.  \color{black}Also, an equivalent lightweight reconstruction model can be developed through knowledge distillation technique.

\small
\textbf{Acknowledgments}: The authors thank SERB, DST, Govt, of India for partially supporting this work through grant no. CRG/2020/005465.
\vspace{-2mm}

\scriptsize
\bibliographystyle{IEEEtran.bst}
\bibliography{mybibfile.bib}

\begin{thebibliography}{10}
\providecommand{\url}[1]{#1}
\csname url@samestyle\endcsname
\providecommand{\newblock}{\relax}
\providecommand{\bibinfo}[2]{#2}
\providecommand{\BIBentrySTDinterwordspacing}{\spaceskip=0pt\relax}
\providecommand{\BIBentryALTinterwordstretchfactor}{4}
\providecommand{\BIBentryALTinterwordspacing}{\spaceskip=\fontdimen2\font plus
\BIBentryALTinterwordstretchfactor\fontdimen3\font minus
  \fontdimen4\font\relax}
\providecommand{\BIBforeignlanguage}[2]{{%
\expandafter\ifx\csname l@#1\endcsname\relax
\typeout{** WARNING: IEEEtran.bst: No hyphenation pattern has been}%
\typeout{** loaded for the language `#1'. Using the pattern for}%
\typeout{** the default language instead.}%
\else
\language=\csname l@#1\endcsname
\fi
#2}}
\providecommand{\BIBdecl}{\relax}
\BIBdecl

\bibitem{1561189}
{Han, J. and Bhanu, Bir}, ``{Individual Recognition Using Gait Energy Image},''
  \emph{{IEEE Trans. on Pattern Analysis and Machine Intelligence}}, vol.~28,
  no.~2, pp. 316--322, 2005.

\bibitem{S0165168411003392}
A.~Roy, S.~Sural, and J.~Mukherjee, ``{Gait Recognition Using Pose Kinematics
  and Pose Energy Image},'' \emph{Signal Processing}, vol.~92, no.~3, pp.
  780--792, 2012.

\bibitem{gupta2021exploiting}
S.~K. Gupta and P.~Chattopadhyay, ``{Exploiting Pose Dynamics for Human
  Recognition from their Gait Signatures},'' \emph{Multimedia Tools and
  Applications}, vol.~80, no.~28, pp. 35\,903--35\,921, 2021.

\bibitem{gupta2021gait}
S.~{K. Gupta} and P.~Chattopadhyay, ``{Gait Recognition in the Presence of
  Co-Variate Conditions},'' \emph{Neurocomputing}, vol. 454, pp. 76--87, 2021.

\bibitem{279851325}
A.~Roy, S.~Sural, J.~Mukherjee, and G.~Rigoll, ``{Occlusion Detection and Gait
  Silhouette Reconstruction from Degraded Scenes},'' \emph{Signal, Image and
  Video Processing}, vol.~5, no.~4, 2011.

\bibitem{Chattopadhyay2014Jan}
P.~Chattopadhyay, A.~Roy, S.~Sural, and J.~Mukhopadhyay, ``{Pose Depth Volume
  Extraction from RGB-D Streams for Frontal Gait Recognition},'' \emph{Journal
  of Visual Communication and Image Representation}, vol.~25, no.~1, pp.
  53--63, 2014.

\bibitem{Chattopadhyay2015Oct}
P.~Chattopadhyay, S.~Sural, and J.~Mukherjee, ``{Frontal Gait Recognition from
  Occluded Scenes},'' \emph{Pattern Recognition Letters}, vol.~63, pp. 9--15,
  2015.

\bibitem{Babaee2018GaitRF}
M.~Babaee, L.~Li, and G.~Rigoll, ``{Gait Recognition from Incomplete Gait
  Cycle},'' in \emph{Proc. of the Intl. Conf. on Image Processing}.\hskip 1em
  plus 0.5em minus 0.4em\relax IEEE, 2018, pp. 768--772.

\bibitem{babaee2018person}
{M. Babaee}, L.~Li, and G.~Rigoll, ``{Person Identification from Partial Gait
  Cycle Using Fully Convolutional Neural Network},'' \emph{Neurocomputing},
  vol. 338, no.~1, pp. 116--125, 2019.

\bibitem{Zhang2010Jul}
{Zhang, Erhu and Zhao, Yongwei and Xiong, Wei}, ``{Active Energy Image Plus
  2DLPP for Gait Recognition},'' \emph{{Signal Processing}}, vol.~90, no.~7,
  pp. 2295--2302, 2010.

\bibitem{8835194}
P.~P. Min, S.~Sayeed, and T.~S. Ong, ``{Gait Recognition Using Deep
  Convolutional Features},'' in \emph{Proc. of the $7^{th}$ Intl. Conf. on
  Information and Communication Technology}.\hskip 1em plus 0.5em minus
  0.4em\relax IEEE, 2019, pp. 1--5.

\bibitem{Shiraga2016Jun}
K.~Shiraga, Y.~Makihara, D.~Muramatsu, T.~Echigo, and Y.~Yagi, ``{GEINet:
  View-Invariant Gait Recognition Using a Convolutional Neural Network},'' in
  \emph{{Proc. of the Intl. Conf. on Biometrics}}, 2016, pp. 1--8.

\bibitem{Alotaibi2017Nov}
M.~Alotaibi and A.~Mahmood, ``{Improved Gait Recognition Based on Specialized
  Deep Convolutional Neural Network},'' \emph{Computer Vision and Image
  Understanding}, vol. 164, pp. 103--110, 2017.

\bibitem{Battistone2019Sep}
F.~Battistone and A.~Petrosino, ``{TGLSTM: A Time Based Graph Deep Learning
  Approach to Gait Recognition},'' \emph{Pattern Recognition Letters}, vol.
  126, pp. 132--138, 2019.

\bibitem{8616800}
X.~Ben, C.~Gong, P.~Zhang, R.~Yan, Q.~Wu, and W.~Meng, ``{Coupled Bilinear
  Discriminant Projection for Cross-View Gait Recognition},'' \emph{IEEE Trans.
  on Circuits and Systems for Video Technology}, vol.~30, no.~3, pp. 734--747,
  2020.

\bibitem{chao2019gaitset}
H.~Chao, Y.~He, J.~Zhang, and J.~Feng, ``{Gaitset: Regarding Gait as a Set for
  Cross-View Gait Recognition},'' in \emph{Proc. of the AAAI Conf. on
  Artificial Intelligence}, vol.~33, no.~01, 2019, pp. 8126--8133.

\bibitem{chao2021gaitset}
H.~Chao, K.~Wang, Y.~He, J.~Zhang, and J.~Feng, ``{GaitSet: Cross-view Gait
  Recognition through Utilizing Gait as a Deep Set},'' \emph{IEEE Trans. on
  Pattern Analysis and Machine Intelligence}, 2021.

\bibitem{fan2020gaitpart}
C.~Fan, Y.~Peng, C.~Cao, X.~Liu, S.~Hou, J.~Chi, Y.~Huang, Q.~Li, and Z.~He,
  ``{Gaitpart: Temporal Part-Based Model for Gait Recognition},'' in
  \emph{Proc. of the IEEE/CVF Conf. on Computer Vision and Pattern
  Recognition}, 2020, pp. 14\,225--14\,233.

\bibitem{song2022distilled}
X.~Song, Y.~Huang, C.~Shan, J.~Wang, and Y.~Chen, ``{Distilled Light GaitSet:
  Towards Scalable Gait Recognition},'' \emph{Pattern Recognition Letters},
  2022.

\bibitem{yu2017gaitgan}
S.~Yu, H.~Chen, E.~B. Garcia~Reyes, and N.~Poh, ``{GaitGAN: Invariant Gait
  Feature Extraction Using Generative Adversarial Networks},'' in \emph{Proc.
  of the Conf. on Computer Vision and Pattern Recognition Workshops}, 2017, pp.
  30--37.

\bibitem{yu2019gaitganv2}
S.~Yu, R.~Liao, W.~An, H.~Chen, E.~B. Garc{\'\i}a, Y.~Huang, and N.~Poh,
  ``{GaitGANv2: Invariant Gait Feature Extraction using Generative Adversarial
  Networks},'' \emph{Pattern Recognition}, vol.~87, pp. 179--189, 2019.

\bibitem{he2018multi}
Y.~He, J.~Zhang, H.~Shan, and L.~Wang, ``{Multi-task GANs for View-Specific
  Feature Learning in Gait Recognition},'' \emph{IEEE Trans. on Information
  Forensics and Security}, vol.~14, no.~1, pp. 102--113, 2018.

\bibitem{zhang2019vn}
P.~Zhang, Q.~Wu, and J.~Xu, ``{VN-GAN: Identity-Preserved Variation Normalizing
  GAN for Gait Recognition},'' in \emph{Proc. of the Intl. Joint Conf. on
  Neural Networks}.\hskip 1em plus 0.5em minus 0.4em\relax IEEE, 2019, pp.
  1--8.

\bibitem{zhang2019vt}
{P.~Zhang}, Q.~Wu, and J.~Xu, ``{VT-GAN: View Transformation GAN for Gait
  Recognition Across Views},'' in \emph{Proc. of the Intl. Joint Conf. on
  Neural Networks}.\hskip 1em plus 0.5em minus 0.4em\relax IEEE, 2019, pp.
  1--8.

\bibitem{han2022unified}
F.~Han, X.~Li, J.~Zhao, and F.~Shen, ``{A Unified Perspective of
  Classification-Based Loss and Distance-Based Loss for Cross-View Gait
  Recognition},'' \emph{Pattern Recognition}, p. 108519, 2022.

\bibitem{chattopadhyay2014frontal}
P.~Chattopadhyay, S.~Sural, and J.~Mukherjee, ``{Frontal Gait Recognition from
  Incomplete Sequences Using RGB-D Camera},'' \emph{IEEE Trans. on Information
  Forensics and Security}, vol.~9, no.~11, pp. 1843--1856, 2014.

\bibitem{zheng2011robust}
S.~Zheng, J.~Zhang, K.~Huang, R.~He, and T.~Tan, ``{Robust View Transformation
  Model for Gait Recognition},'' in \emph{Proc. of the Intl. Conf. on Image
  Processing}.\hskip 1em plus 0.5em minus 0.4em\relax IEEE, 2011, pp.
  2073--2076.

\bibitem{6215042}
H.~Iwama, M.~Okumura, Y.~Makihara, and Y.~Yagi, ``{The OU-ISIR Gait Database
  Comprising the Large Population Dataset and Performance Evaluation of Gait
  Recognition},'' \emph{IEEE Trans. on Information Forensics and Security},
  vol.~7, no.~5, pp. 1511 -- 1521, 2017.

\bibitem{hofmann2011gait}
M.~Hofmann, S.~Sural, and G.~Rigoll, ``{Gait Recognition in the Presence of
  Occlusion: A New Dataset and Baseline Algorithms},'' in \emph{Proc. of the
  $19^{th}$ Intl. Conf. on Computer Graphics, Visualization and Computer
  Vision}, 2011.

\bibitem{das2019rgait}
D.~Das, A.~Agarwal, P.~Chattopadhyay, and L.~Wang, ``{RGait-NET: An Effective
  Network for Recovering Missing Information from Occluded Gait Cycles},''
  \emph{arXiv preprint arXiv:1912.06765}, 2019.

\bibitem{carass2020evaluating}
A.~Carass, S.~Roy, A.~Gherman, J.~C. Reinhold, A.~Jesson, T.~Arbel, O.~Maier,
  H.~Handels, M.~Ghafoorian, B.~Platel \emph{et~al.}, ``{Evaluating White
  Matter Lesion Segmentations with Refined S{\O}Rensen-Dice Analysis},''
  \emph{Scientific Reports}, vol.~10, no.~1, pp. 1--19, 2020.

\bibitem{royocclusion}
A.~Roy, S.~Sural, J.~Mukherjee, and G.~Rigoll, ``{Occlusion Detection and Gait
  Silhouette Reconstruction from Degraded Scenes},'' \emph{Signal, Image and
  Video Processing}, vol.~5, no.~4, pp. 415-- 430, 2011.

\bibitem{article1}
{Chen, Changhong and Liang, Jimin and Zhao, Heng and Hu, Haihong and Tian,
  Jie}, ``{Frame Difference Energy Image for Gait Recognition with Incomplete
  Silhouettes},'' \emph{{Pattern Recognition Letters}}, vol.~30, no.~11, pp.
  977--984, 08 2009.

\end{thebibliography}
\newpage
\normalsize
\textbf{Authors' Response}\\\\
\small
We are thankful to the editors and the reviewers for their efforts in reading the manuscript titled “Gait Cycle Reconstruction and Human Identification from Occluded Sequences” thoroughly and providing us with constructive feedback and suggestions to improve the quality of the manuscript further. In the revised manuscript, we have sincerely handled the concerns of the individual reviewers and have been able to come up with an improved version of the previously submitted manuscript. Additionally, we have made efforts to keep the discussions crisp and concise so that all the results and discussions can be fit within the given page limit, remove some old references and add a few recent references and correct a few grammatical errors present in the previously submitted version of the manuscript. Specific responses to the reviewers’ comments are given below. Major changes made in the text have been highlighted in red fonts.\\\\
\textbf{Associate Editor}\\\\
\color{blue}
The motivation of this paper should be further clarified. CASIA-B and OU-ISIR datasets are not challenging. I suggest that some wild-word gait data should be acquired for the experiments. \\\\
\color{black}
\textbf{Response}: In response to the first suggestion from the Associate Editor, we have clarified the motivation behind the work in further details in the Introduction Section, i.e., Section I of the revised manuscript. Please refer to the updated text in the first paragraph of Section I in Page 1 of the revised manuscript. In response to the second suggestion from the Associate Editor, we have conducted similar experiments with an additional dataset, namely the TUM-IITKGP dataset, which is the only gait dataset in the public domain with real occluded sequences. Our model has also given quite satisfactory results on this dataset, as shown in Fig. 6(b) (Page 6) and Table III (Page 7). However, in the work we have primarily focused on occlusion reconstruction from the fronto-parallel view (i.e., side view) gait sequences and trained the proposed occlusion reconstruction model by constructing an extensive gallery set of side view sequences. However, our model would be equally effective in reconstructing sequences from any other view since the sub-networks involved in the proposed model basically utilize the spatio-temporal information from the available frames to reconstruct the missing/occluded frame. \\\\

\textbf{Reviewer 1}\\\\
\color{blue}
\textbf{General Comments}: This manuscript proposes a deep learning method to reconstruct missing frames in a video clip of gait. The manuscript has presented well.\\\\ \color{black}
\textbf{Response}: We thank the reviewer for his/her encouraging comments.\\\\
\color{blue}
\textbf{Suggestions for Improvements}:\\ 
\textbf{Comment 1}: My major criticism is that the comparison with other methods in the literature has been presented in the form of recognition rate measurements. However a better comparison measurement would be the similarity of the frames reconstructed by the proposed method and other methods in the literature to the ground truth.\\\\
\color{black}
\textbf{Response}: We thank the reviewer for pointing this out. Since some of the existing approaches carry out GEI feature reconstruction instead of frame-level reconstruction, instead of computing the similarity between the reconstructed and ground truth frames we compute the similarity between the GEIs derived from the frames reconstructed by each occlusion reconstruction method with that derived from the ground truth. Please refer to Table 2 in Page 6, Column 2 of the revised manuscript for the comparative results. \\\\
\textbf{Reviewer 2}\\\\
\color{blue}General Comments: The paper proposes an effective method to reconstruct the occluded frames before carrying out gait recognition. The proposed method exploits LSTM networks to predict an embedding of each occluded/corrupted frame from the forward and backward directions. Next, fuse the predictions of the two LSTMs through a network of residual blocks and convolutional layers. The paper well addressed the problem and generated interesting results in such a problem.\\\\
\color{black}
\textbf{Response}: We thank the reviewer for his/her encouraging comments.\\\\
\color{blue}Suggestions for Improvements: To make the article suitable to be published, I suggest the authors revise the paper to address the following concerns.\\\\
Comment 1: The paper's motivation is not clear. Why do we need to solve this problem? \\\\
\color{black}
\textbf{Response}: As per the reviewer’s suggestion we have included the motivation behind the work in the Introduction section of the manuscript. Please refer to the first paragraph of Section I in Page 1 of the revised manuscript.\\\\
\color{blue}
\textbf{Comment 2}: Synthetic occlusions are not convincing in real life. What happens when partial occlusions occur? (half-body occlusions)\\\\
\color{black}
\textbf{Response}: We appreciate the reviewer’s concern over this matter. The effectiveness of our proposed model is independent of whether an input sequence contains partial or full-body occlusions since while reconstructing an occluded frame the forward LSTM utilizes the spatio-temporal information from a set of previous unoccluded frames excluding the frame itself, whereas the backward LSTM exploits the spatio-temporal information from a set of subsequent unoccluded excluding the frame itself. Finally, these two predictions are fused together using a set of residual blocks and convolutional layers to come up with a refined prediction about the occluded frame. It may be noted that the occluded frame is not used for making either the forward or the backward predictions. We have mentioned this fact in Page 3 in the first paragraph of the Proposed Approach Section, i.e., Section III and also in Page 8 in the Conclusions and Future Work Section, i.e., Section V. In the revised manuscript, we have shown results after testing our algorithm using real occluded sequences of the TUM-IITKGP dataset (which consists of partial occluded sequences as well) and have achieved quite high recognition accuracy. Please refer to Fig. 6(b) in Page 6, Column 2 and Table III in Page 7, Column 1 of the revised manuscript.\\\\
\color{blue}
\textbf{Comment 3}: The reviewer has checked both datasets, the CASIA-B and the OU-ISIR. In my opinion, those datasets are not challenging at all. Uniform lighting conditions, uniform background, high contrast participants, and participants only cover two paths (left to right, right to left). A real-environment dataset must be tested.\\\\
\color{black}
\textbf{Response}: We thank the reviewer for his/her critical comment. There exists only one gait data set in the public domain that contains sequences with real partially/fully occluded frames, namely the TUM-IITKGP dataset. The dataset contains videos of real static and dynamic occlusions from 35 individuals from the fronto-parallel view, i.e., side-view walking either from left to right or right to left. In the present work, we have focused solely on developing a neural network-based occlusion reconstruction algorithm suitable for predicting missing binary frames in side view sequences and to the best of our knowledge TUM-IITKGP data is the only dataset to test our model’s effectiveness. Through experiments, we have observed that none of the other occlusion reconstruction algorithms in gait recognition reconstructs frames or GEIs as effectively as ours.  Also, our proposed model can be suitably trained to reconstruct occluded sequences from any other view as long as some temporal information can be extracted from the sequence. Fusing an effective view transformation model with an occlusion reconstruction model can help in performing gait recognition in a real environment where the scene is likely to contain occlusion with multiple moving persons in varying directions. This is still an open area of research which we would like to take up next, as also stated in Section V of the revised manuscript.\\\\
\color{blue}
\textbf{Comment 4}: Fig 2 must be improved. I can not read anything.\\\\
\color{black}
\textbf{Response}: As per the reviewer’s suggestion, we have improved the clarity of Fig. 2 by rotating it and presenting it at a higher scale. Please refer to Page 3, Column 2 of the revised manuscript.\\\\
\color{blue}
\textbf{Comment 5}: Do you mix both datasets during training? Please, train and test with different datasets. You may be biased if participants appear in several clips.\\\\
\color{black}
\textbf{Response}: We thank the reviewer for pointing out this lack of explanation in the previously submitted version of the manuscript. In the revised manuscript we have clarified the fact that training of the reconstruction model is done using an extensive dataset formed using samples from CASIA-B and OU-ISIR data sets, whereas the recognition model is trained using sequences specific to a particular data set at a time. Please refer to Page 5, Column 1 of the revised manuscript for the related explanation. Further, in our experiments, none of the samples of the test set has been considered as part of the gallery set for training either the reconstruction or the recognition models. We have clarified this fact clearly in Page 5, Columns 1-2 under the heading Experiments and Results of the revised version of the manuscript.\\\\ 
\color{blue}
\textbf{Comment 6}: Some references are ten to twenty years old. Please update SOTA.\\\\
\color{black}
\textbf{Response}: We thank the reviewer for his/her suggestion. Some of the old references stated in the paper are foundational work in Computer Vision-based gait recognition and these references have been retained in the revised manuscript as well. But we removed the ones that are quite old and also not regarded as foundational work in this domain. Additionally, we have added a few recent references, namely citation numbers [18], [19], and [25] to make the SOTA up-to-date, and also included discussions related to these papers in the Related Work section, i.e., Section II. Please refer to Page 2 of the revised manuscript.\\\\

\textbf{Reviewer 3}\\\\
\color{blue}
\textbf{General Comments}: This paper proposes an effective method to reconstruct the occluded frames before carrying out gait recognition. The experimental results verify the effectiveness of the method. The paper is well written and the description is clear enough to reproduce the proposed method. The authors have conducted evaluations on gait recognition based on different recognition methods. The results show that reconstruction before recognition brought a large performance increase.\\\\
\color{black}
\textbf{Response}: We thank the reviewer for his/her encouraging comments.\\\\
\color{blue}
\textbf{Suggestions for Improvements}:\\
\textbf{Comment 1}: Ablation study is missing to check the roles of each part of the proposed method.\\\\
\color{black}
\textbf{Response}: As per the reviewer’s suggestion, in the revised manuscript, we have performed an ablation study to check the rank-based classification accuracy of gait recognition of each of the forward LSTM, the backward LSTM, and the proposed fusion model. Corresponding results on the synthetically occluded sequences of the CASIA-B data are shown in Figs. 7(a)-(c), Page 7, Column 2 of the revised manuscript. The results reveal that the proposed fused reconstruction model is superior to either of the LSTMs and has thus resulted in achieving a higher recognition accuracy at each of the different ranks and varying degrees of occlusion than that of the individual LSTMs.\\\\
\color{blue}
\textbf{Comment 2}: It is better to add some comparisons on the reconstruction quality to show the merits of the proposed method.\\\\
\color{black}
\textbf{Response}: We thank the reviewer for his/her constructive suggestion. In the revised manuscript, we have included comparisons on the reconstruction quality of the different methods by computing the similarity between the GEIs from the reconstructed sequence and GEIs from the ground truth unoccluded sequence using the Dice score metric. Please refer to Table II, Page 6, Column 2 of the revised manuscript for these comparative results. Additionally, in Table I (Page 6, Column 1), we have stated Dice scores along with the recognition accuracy for each of the varying degrees of synthetic occlusion applied on the CASIA-B and the OU-ISIR LP datasets.

\end{document}